\documentclass[cameraready]{Interspeech}
\usepackage{tipa}

\title{A Speech Corpus for Mizo Automatic Speech Recognition: Whisper and SraVaani 1.0 Fine-Tuning with Morphology-Aware Evaluation}

\author[affiliation={}, orcid=0000-0002-9051-1255]{Priyankoo}{Sarmah}

\author[affiliation={}, orcid=0000-0002-9051-1255]{Sanasam Ranbir}{Singh}

\author[affiliation={},orcid=0000-0001-9069-9748]{Lalhmingmawia}{}

\address{
    Center for Linguistic Science and Technology\\
    Indian Institute of Technology Guwahati, India
}

\email{\{priyankoo,ranbir\}@iitg.ac.in, mapuia442@gmail.com}

\keywords{ASR, Mizo, Whisper fine tuning,SraVaani 1.0}

\newcommand{\blue}[1]{\textcolor{blue}{#1}}

\usepackage{comment}

\begin{document}
\maketitle
\begin{abstract}
This study reports the development of an Automatic Speech Recognition (ASR) system in Mizo, a low-resource language. The development included collecting $17.62$ hours of speech data, curating it, and fine-tuning the Mizo ASR system with three Whisper multilingual models and with the SraVaani $1.0$ Indic multilingual model. Whisper-large-v3 achieved the lowest conventional WER ($18.08$\%), while morphology-aware evaluation yielded a WER of $7.22$\%. Zero-shot evaluation of the SraVaani 1.0 Indic multilingual model yielded a WER of 58.27\%, while Mizo-specific fine-tuning reduced the conventional WER to 29.45\% and the morphology-aware WER to 17.93\%. The results demonstrate that the Whisper model can achieve a substantially low WER, even when adapted to an unseen language. In contrast, SraVaani $1.0$ supports the Mizo language in its multilingual model; however, fine-tuning with carefully curated Mizo speech data substantially improves its performance.

\end{abstract}

\section{Introduction}
Mizo is a low-resource Tibeto-Burman language spoken in the Mizoram province of India, and in parts of Myanmar and Bangladesh. This paper reports the development of a spoken data resource for Mizo Automatic Speech Recognition (ASR) system and the subsequent fine-tuning of publicly available Mizo ASR models to build an ASR system for the language. Specifically, two multilingual models are fine-tuned on our data: Whisper \cite{radford2023robust} and SraVaani 1.0 \cite{pulikodan2026sravaani}. Mizo is not included in Whisper's original language set, whereas it is supported by SraVaani 1.0. We therefore investigate whether pretrained multilingual Whisper models can be effectively adapted to Mizo using the available training data.

It has been reported that Whisper models can be successfully adapted to previously unseen languages \cite{unseen1}. A study on Kildin-Sami showed that fine-tuning with only 30 minutes of transcribed Kildin-Sami speech using the Russian-based Whisper model yielded a modest WER of 68.55\% \cite{unseen-sami}. These findings motivate the investigation of Whisper-based adaptation to Mizo. In contrast, SraVaani 1.0 \cite{pulikodan2026sravaani} supports Mizo; hence, our first goal is to conduct a zero-shot test on our Mizo test set using the SraVaani 1.0 multilingual model. The model was subsequently fine-tuned on the Mizo training data to assess the effect of language-specific adaptation.

The development of Mizo continuous ASR has progressed in stages over the last $10$ years. Starting with simple phone and digit recognition systems for Mizo \cite{mizo-phone,mizo-digit}, it has progressed to domain-specific, limited-vocabulary speech recognition and continuous speech recognition \cite{mizo-agri, mizo-cont}. Recently, studies have shown the effectiveness of fine-tuning with Wav2vec 2.0 and XLS-R models, resulting in substantially improved performance in Mizo ASR \cite{mizo-andrew}. It is reported that the Wav2vec-Base-Mizo-Lus and the XLS-R-300M-Mizo-Lus models' WERs are 16.59\% and 11.84\%, respectively \cite{mizo-andrew}. 

A major limitation in the development of Mizo ASR systems has been the limited availability of publicly accessible speech data. Hence, in this work, we report the collection and curation of speech data, which is released to support further development of Mizo language technologies. The development of the corpus is reported in Section \ref{sec:database}. Following that, the Mizo speech data was used to fine-tune the Whisper-small, Whisper-medium, and Whisper-large-v3 multilingual models to adapt to the Mizo speech. The performance of the three systems is reported in Section \ref{sec:results}. In addition to Whisper models, we have fine-tuned the SraVaani 1.0 model using Mizo data from our corpus. We evaluated the three fine-tuned Whisper models, the SraVaani 1.0 multilingual model, and the SraVaani 1.0 Mizo fine-tuned model. The results of the evaluations are reported in Section \ref{sec:results}.

\section{Mizo Speech corpus}
\label{sec:database}

\subsection{Data collection}
The Mizo speech corpus was remotely collected over a period of two months. A list of approximately 8,000 Mizo sentences was prepared and validated by a native Mizo speaker. The domain of the text comprised newspaper items and court judgments, translated from English into Mizo by native speakers of the language. The sentences were presented to the Mizo speakers in their native writing system through a web interface. The web interface's recording function allowed participants to record their reading of the displayed sentence. Participants subsequently submitted the recordings through the web interface.

\begin{figure}[]
  \centering
  \includegraphics[width=\linewidth]{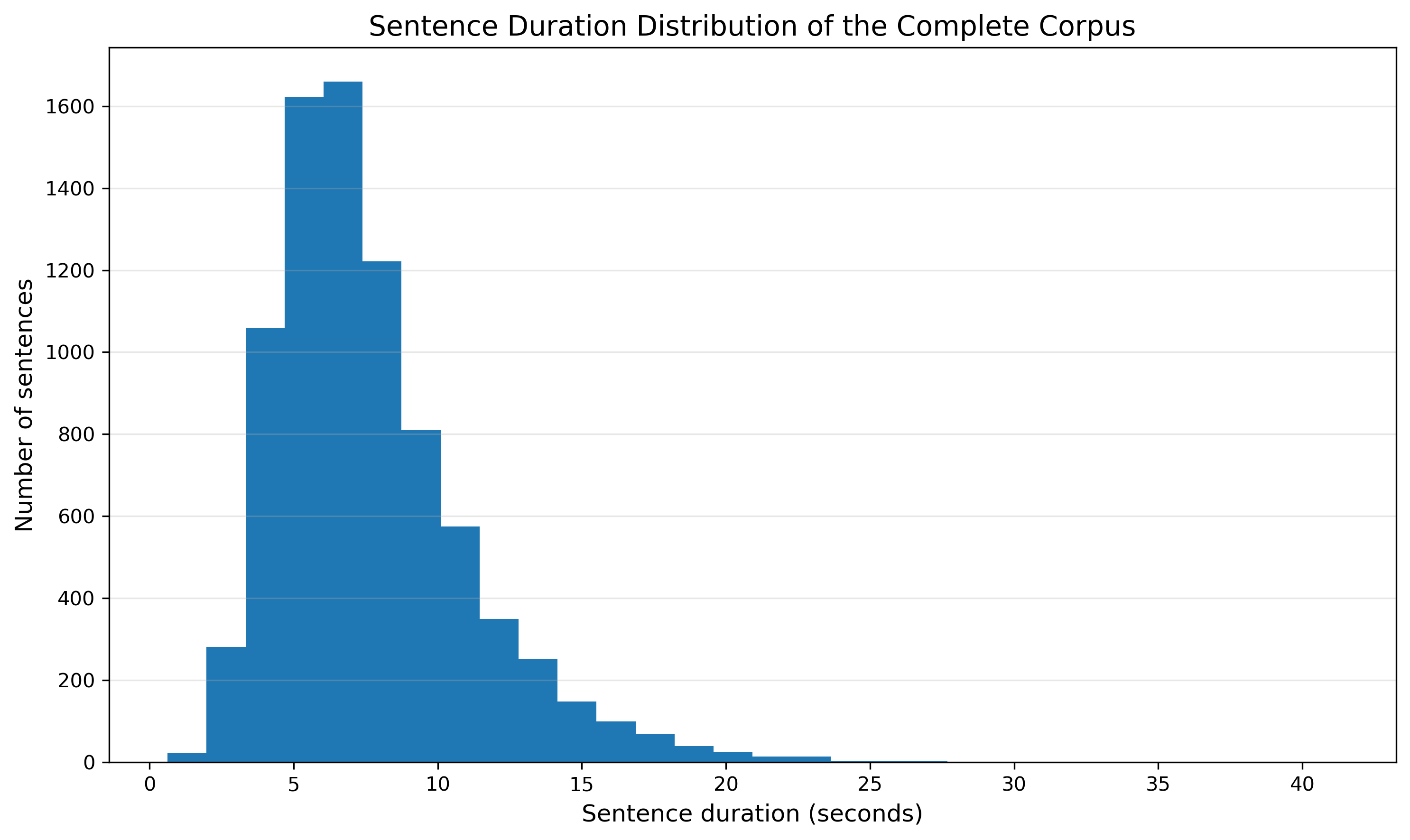}
  \caption{Histogram of duration of speech files in the corpus.}
  \label{fig:hist}
\end{figure}

\begin{table}[]
  \caption{Data split for training, validation and testing.}
  \label{tab:word_styles}
  \centering
  \begin{tabular}{lccc}
    \toprule
   & \textbf{Speakers}&\textbf{Sentences}&\textbf{Hours}  \\
    \midrule
Training&184&7656&16.18\\
Validation&11&426&1.02\\
Testing&05&192&0.42\\
    \bottomrule
  \end{tabular}
\end{table}

\begin{table}[]
  \caption{Durational characteristics of the speech database}
  \label{tab:corpus_stats}
  \centering
  \begin{tabular}{lr}
    \toprule
 
Sentences&8274\\
Total duration&17.62 hours\\
Minimum duration&0.63 seconds\\
Maximum duration&41.22 seconds\\
Mean duration&7.67 seconds\\
Median duration&6.94 seconds\\

    \bottomrule
  \end{tabular}
\end{table}

In total, 200 Mizo speakers participated in recording the data. The recording interface allowed each participant to record a maximum of $50$ sentences. However, after the participants submitted their recordings, some were corrupted, and some files contained no speech. Hence, after both automatic and manual filtering, the database consisted of 8274 sentences, as shown in Table \ref{tab:word_styles}. Depending on the device and browser used by the participants, the recordings came in $7$ different formats: .mov, .mp4, .m4a, .3g, .3g2, .mj2, and .webm. However, all recordings were consistently sampled at 48000 Hz, and 90\% were in .webm format.

\subsection{Data curation and preprocessing}
The recordings were saved with the sentence ID and speaker ID information available in the filename. Apart from that, the date of recording was also encoded in the filename, resulting in a filename as SPYYYYMMDD\_ABC123\_MZ\_NXXXX\_FORM.wav, where YYYYMMDD records the date of speech recording, ABC123 is an alphanumeric code for the speaker, NXXXX is the sentence number identifying the corresponding text, and FORM is one of the seven formats in which the data was recorded.

Once the sound files are stored, they are converted into .wav format with a sampling rate of 16000 Hz for further processing. During conversion, any corrupted files were removed from the database, resulting in a speech corpus of 8698 sound files. A native Mizo speaker subsequently reviewed all 8,698 recordings to verify correspondence between the speech and reference transcripts. Apart from that, recordings containing only background noise or incomplete speech were discarded. Finally, a database of 8274 .wav files, each containing a sentence, was constructed. The average sentence duration in the database was about 8 seconds. Table \ref{tab:corpus_stats} provides statistics of the recorded speech database in the current study. Figure \ref{fig:hist} shows the distribution of sentence duration in the corpus. 

\subsection{Corpus partition}
The speech database of 8274 .wav files was divided into training, validation, and test sets, as shown in Table \ref{tab:word_styles}, ensuring that no speakers overlap across the three sets. The same speaker-independent training, validation, and test sets were used to fine-tune the Whisper-small, Whisper-medium, Whisper-large-V3, and SraVaani 1.0 models.

\section{Methodology}
\label{sec:method}
\subsection{Overview of the experimental framework}

\begin{figure*}[t]
  \centering
  \includegraphics[width=\linewidth]{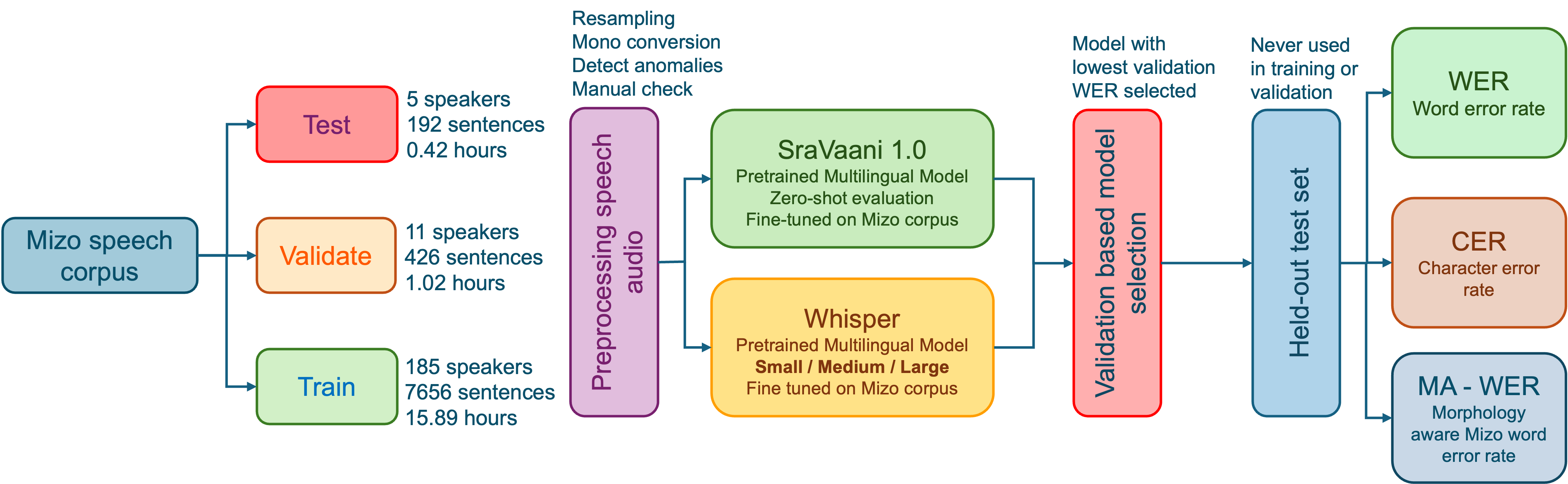}
  \caption{Schematic diagram showing the flow of the experimental protocol.}
  \label{fig:flow}
\end{figure*}

The experimental framework comprised four stages: model adaptation, validation-based model selection, held-out test evaluation, and language-specific error analysis. The collected speech data was manually validated for consistency with the text that the speakers produced. Once that is confirmed, the data was downsampled to 16000 Hz, as required by the models for fine-tuning. Secondly, the pretrained multilingual ASR models were adapted for Mizo. Thirdly, model selection was performed based on the lowest WER achieved during validation. Fourthly, a held-out test set comprising only Mizo speakers was used to evaluate the Mizo ASRs. Given the morphological and orthographic characteristics of Mizo, a Mizo-specific WER calculation metric was introduced in the study. The curated Mizo corpus of $17.62$ hours was used to adapt three multilingual Whisper models and SraVaani $1.0$. All systems were trained and evaluated using the same speaker-independent data partitions. Model selection was performed on the validation set, while the held-out test set was used exclusively for final evaluation. The entire methodological flow is shown in Figure \ref{fig:flow}.

All model fine-tuning and inference experiments were conducted on a workstation equipped with an NVIDIA RTX PRO 4500 Blackwell GPU with 32~GB of VRAM. The experiments were implemented using Python~3.10, PyTorch, Hugging Face Transformers, and NVIDIA NeMo. Whisper models were fine-tuned using the Hugging Face Transformers framework, while SraVaani~1.0 fine-tuning was performed using NVIDIA NeMo. GPU acceleration through CUDA was used for all training experiments.

\subsection{Whisper fine-tuning}
Whisper is a multilingual encoder-decoder Transformer-based model for automatic speech recognition and speech translation. While the Whisper-small and Whisper-medium models are trained on $680$k hours of data, their parameter counts vary. On the other hand, Whisper-large-v3 is trained on more than $5$ million hours of speech data. The salient features of the three Whisper models used for fine-tuning in the current work are provided in Table \ref{tab:model_characteristics}. As shown in Table \ref{tab:model_characteristics}, the three models employ the same encoder--decoder Transformer-based ASR framework but differ in terms of model capacity. While $99$ languages are supported by the Whisper models, Mizo is not one of them. Hence, Mizo-specific linguistic knowledge is not explicitly represented in the original Whisper language set. These experiments assess the adaptation of an unseen Tibeto-Burman language across three Whisper model scales.

\begin{table*}[ht]
\centering
\caption{Characteristics of the multilingual ASR models evaluated in this study.}
\label{tab:model_characteristics}
\begin{tabular}{lcccc}
\hline
\textbf{Model} & \textbf{Architecture} & \textbf{Parameters} &
\textbf{Pretraining data} & \textbf{Mel bins} \\
\hline
Whisper-small
& Encoder--Decoder Transformer
& 244 M
& 680k h
& 80 \\

Whisper-medium
& Encoder--Decoder Transformer
& 769 M
& 680k h
& 80 \\

Whisper-large-v3
& Encoder--Decoder Transformer
& 1,550 M
& $\sim$5M h
& 128 \\

SraVaani 1.0
& FastConformer Hybrid RNNT/CTC
& $\sim$430 M
& $\sim$31k h
& -- \\
\hline
\end{tabular}
\end{table*}

\label{sec:ASR}

\begin{table*}
\centering
\caption{Fine-tuning configurations used for the ASR models.}
\label{tab:finetuning_config}
\begin{tabular}{lcccccc}
\hline
\textbf{Model} & \textbf{Optimizer} & \textbf{LR} &
\textbf{Effective batch size} & \textbf{Epochs} &
\textbf{Selection} \\
\hline
Whisper-small
& Adafactor
& $5\times10^{-6}$
& 16
& 20
& Best val. WER \\

Whisper-medium
& Adafactor
& $5\times10^{-6}$
& 16
& 20
& Best val. WER \\

Whisper-large-v3
& Adafactor
& $5\times10^{-6}$
& 16
& 20
& Best val. WER \\

SraVaani 1.0
& AdamW
& $1\times10^{-4}$
& 16
& 20 + extended
& Best val. WER \\
\hline
\end{tabular}
\end{table*}

The fine-tuning parameters for the Whisper training are detailed in Table~\ref{tab:finetuning_config}. As shown in the Table, the learning rate, batch size, and number of epochs were fixed during Whisper fine-tuning. For each model, of the 20 epochs, the one that produced the checkpoint with the best validation WER was selected as the model for further testing. 

\subsection{SraVaani 1.0 fine-tuning}
SraVaani 1.0 is a multilingual Indic automatic speech recognition model based on a FastConformer encoder and a hybrid RNNT/CTC architecture. In contrast to Whisper, SraVaani 1.0 supports Mizo among the languages the pretrained model supports. We first evaluated the pretrained SraVaani 1.0 model directly on the
held-out Mizo test set to establish a zero-shot baseline. The model was subsequently fine-tuned using the Mizo training corpus to quantify the effect of Mizo-specific adaptation.

For Mizo adaptation, the pretrained Conformer encoder was frozen, and the decoder and joint components were fine-tuned. This strategy reduced the number of trainable parameters while retaining the pretrained acoustic representations of the multilingual model. The fine-tuning was performed using AdamW with a learning rate of
$1\times10^{-4}$ and a weight decay of $1\times10^{-3}$. A micro-batch size of 1 with gradient accumulation of 16 was used, resulting in an effective batch size of 16, as shown in Table \ref{tab:finetuning_config}. Validation was performed after every training epoch, and the checkpoint achieving the lowest validation WER was selected for final evaluation.

An initial training run was conducted for 20 epochs. While the last two epochs had slightly higher WER, training was extended to 25 epochs to determine whether additional optimization improved validation WER. However, the validation WER did not improve in the additional runs. Hence, the 18\super{th} epoch, which yielded the lowest WER, was selected as the model for testing. 

\subsection{Model selection and evaluation}
Validation WER was computed after each training epoch, and the checkpoint with the lowest validation WER was selected for final test evaluation. Table \ref{tab:ft-epoch} shows the epoch corresponding to the best WER for all four fine-tuned models. While all the Whisper models were fine-tuned up to the 20\super{th} epoch, for the SraVaani 1.0 model, fine-tuning was continued until the 25\super{th} epoch; however, as seen in Table \ref{tab:ft-epoch} the 18\super{th} epoch yielded the lowest WER, which was selected for further testing. 

\begin{table}
\centering
\caption{Best epochs and corresponding validation WER of four fine-tuned models.}
\label{tab:ft-epoch}
\begin{tabular}{lcc}
\hline
\textbf{Model} & \textbf{Best epoch} & \textbf{Validation WER}\\
\hline
Whisper-small Mizo{--}FT&$15$&$28.99$\\
Whisper-medium Mizo{--}FT&$13$&$26.51$\\
Whisper-large-v3 Mizo{--}FT&$13$&$23.00$\\
SraVaani 1.0 Mizo{--}FT&$18$&$33.81$\\
\hline
\end{tabular}
\end{table}

The performance of the ASR systems was evaluated using Word Error Rate (WER), Character Error Rate (CER), and a Mizo-specific morphology-aware WER. All metrics were computed on the held-out test set after applying the same text normalization procedure to both reference transcripts and ASR hypotheses. The normalization included lowercasing, removal of selected punctuation marks, normalization of whitespace, and treating the underscore character as equivalent to the Mizo character \textit{ṭ}.

\subsubsection{Word Error Rate}

Word Error Rate (WER) is the conventional metric used to evaluate automatic speech recognition systems and is defined as

\[WER = \frac{S + D + I}{N},\]

where $S$, $D$, and $I$ denote the number of substitutions, deletions, and insertions, respectively, and $N$ is the number of words in the reference transcription. Lower WER values indicate
better recognition performance.

\subsubsection{Character Error Rate}

Character Error Rate (CER) was additionally computed to measure recognition performance at the character level. CER is calculated in the same manner as WER, but the basic units are characters rather than words:

\[CER = \frac{S_c + D_c + I_c}{N_c},\]

where $S_c$, $D_c$, and $I_c$ represent character substitutions, deletions, and insertions, respectively, and $N_c$ is the number of characters in the reference transcription. CER is particularly useful for Mizo because recognition differences occur at the word-boundary and orthographic levels.

\subsubsection{Morphology-Aware Mizo WER}

Standard WER can overestimate recognition errors in Mizo when the recognized character sequence is correct but the placement of whitespace differs from that of the reference transcription. Such differences can arise from variation in the representation of morphological boundaries in written Mizo. A substantial proportion of the observed word-level discrepancies is associated with differences in whitespace segmentation in Mizo. The underlying reason lies in the variation in morphological boundary marking in Mizo, where a representation such as [kanin] with the morphological marker [in] may optionally be written as [kan in]. For instance, the lexeme [in] is usually written together as a prefix [inentir] and suffix [lungin]. It remains distinct when used as a noun [kan in], a verb [ka in] and a pronoun [in vaiin]. Such spacing irregularities are common across different domains of language use. Even though such forms are considered incorrect in the academic domain, they do appear in everyday writings and readers usually have no trouble interpreting them. Hence, both forms are accepted.

To account for these differences, we introduced a morphology-aware Mizo WER. The metric first tokenizes the reference and hypothesis transcriptions into whitespace-delimited tokens. In addition to the standard one-to-one token comparison, it permits concatenating up to four adjacent reference tokens and up to four adjacent hypothesis tokens. A grouped reference-hypothesis pair is counted as a correct match only when their concatenated character sequences are exactly identical. No approximate, phonetic, or spelling-based matching is performed. For example, a reference sequence
\[\text{a ni}\]
and a hypothesis sequence
\[\text {ani}\]
are considered equivalent when
\[\text{a}+\text{ni}=\text{ani}\]

Insertions and deletions are retained in the standard manner, while the resulting edit distance is normalized by the number of tokens in the reference transcription. The resulting metric is reported as morphology-aware Mizo WER.

\section{Results}        
\label{sec:results}
\subsection{Overall ASR performance}

The results of the evaluation of the three fine-tuned Whisper models, SraVaani 1.0 base model, and SraVaani 1.0 Mizo fine-tuned model, are provided in Table \ref{tab:final_results}. As shown in Table \ref{tab:final_results}, among all the models, Whisper-large-v3 performs the best, with the lowest raw WER of $18.08\%$. Under morphology-aware evaluation, the corresponding MA-WER is $7.22\%$. In fact, across all models, MA-WER is much lower than raw WER, indicating that a considerable portion of errors stems from variation in morphological boundaries.

The zero-shot WER obtained on our held-out Mizo test ($58\%$) set was substantially higher than the Mizo WER reported in the original SraVaani evaluation \cite{pulikodan2026sravaani}. This indicates that conventional WER overestimates the error rate, in part due to mismatches in the placement of morphological boundaries. Hence, MA-WER score for the SraVaani 1.0 is $36.27\%$. Fine-tuning the SraVaani 1.0 multilingual model with our Mizo data substantially improved the results. The raw WER of the Mizo fine-tuned  SraVaani 1.0 is $29.45\%$, and the MA-WER is $17.93\%$. Considering SraVaani has released only the initial version, it shows potential to become a much more robust low-resource Indian language model.

\subsection{Effect of Model Capacity and Mizo Adaptation}
Performance improved consistently as Whisper model capacity increased, as shown in Table \ref{tab:final_results}. This closely mimics the pattern observed in the validation WER reported in Table \ref{tab:ft-epoch}. In terms of MA-WER, the Whisper-medium model improved by $22\%$ from the Whisper-small model. Whisper-large-v3 reduced MA-WER by approximately 19\% relative to Whisper-medium. On the other hand, compared to the base SraVaani 1.0 model, the Mizo fine-tuned model improved by about $51\%$. The results show that the capacity of Whisper models has a significant effect on fine-tuning accuracy. At the same time, fine-tuning the SraVaani 1.0 model yielded improved results in raw WER and MA-WER.

\begin{table*}
\centering
\caption{Results of evaluation on five models.}
\label{tab:final_results}
\begin{tabular}{lccc}
\hline
\textbf{Model} & \textbf{CER (\%)} & \textbf{WER (\%)} &
\textbf{MA-WER (\%)}\\
\hline
Whisper-small Mizo{--}FT&04.83&24.00&11.49\\
Whisper-medium Mizo{--}FT&04.02&21.69&08.87\\
Whisper-large-v3 Mizo{--}FT&03.26&18.08&07.22\\
SraVaani 1.0&17.71&58.27&36.27\\
SraVaani 1.0 Mizo-FT&06.90&29.45&17.93\\
\hline
\end{tabular}
\end{table*}

\begin{table*}
\centering
\caption{Error distribution by model. Foreign-script outputs are reported as the number of utterances; all other entries indicate the number of error occurrences.}
\label{tab:errors}
\begin{tabular}{lcccccc}
\hline
\textbf{Model} & \textbf{Foreign script} & \textbf{Names} &
\textbf{Glottal stops}&\textbf{Numeral transcripts}&\textbf{Code-mix error}&\textbf{$<$\textsubdot{t}$>$}\\
\hline
Whisper-small Mizo{--}FT&NIL&17&7&2&9&0\\
Whisper-medium Mizo{--}FT&NIL&18&6&5&3&0\\
Whisper-large-v3 Mizo{--}FT&4 sentences&8&4&3&2&0\\
SraVaani 1.0&21 sentences &44&9&19&49&31\\
SraVaani 1.0 Mizo{--}FT&NIL&24&6&2&19&4\\
\hline
\end{tabular}
\end{table*}

\subsection{Morphology-Aware Evaluation}
The morphology-aware metric produced consistently lower error rates than conventional WER across all evaluated systems. As shown in Table \ref{tab:final_results}, the results show a substantial reduction from conventional WER to MA-WER across all evaluated systems. The results indicate that a considerable number of errors are due to the incorrect placement of morphological boundaries. Normal Mizo orthography also demonstrates such variation. Although such variation may deviate from formal orthographic conventions, it does not necessarily alter the intended interpretation of the utterance. Using the Mizo morphology-aware evaluation yielded the largest decrease in WER for the fine-tuned Whisper-large-v3 model, while SraVaani~1.0 in the zero-shot condition showed the largest absolute gap between conventional WER and MA-WER. The results show that raw WER may overestimate the errors, whereas the change due to morphological boundary placement may not hinder the intended linguistic content of the sentences. This may be true for several Tibeto-Burman languages using the Roman script.

\subsection{Error Analysis}
The transcriptions produced by the models were manually checked by the third author of this paper, a native speaker of Mizo. Manual inspection identified six major error categories. The analysis was conducted on the complete set of model outputs from the held-out test set, and each error was assigned to one or more of the predefined categories.

\begin{enumerate}
    \item Sentences appearing in non-Mizo script
    \item Names of people and places
    \item Deletion/over-representation of glottal stop /h/
    \item Inconsistent numeral transliteration
    \item Wrong detection of code-mixed English
    \item Misrepresentation of $<$\textsubdot{t}$>$
\end{enumerate}

Apart from that, the morphological boundary marking white space is also inconsistent with the ground truth texts. However, as mentioned earlier, that problem may be attributed to the Mizo script's inconsistency in marking morphological boundaries, which may not hinder a normal Mizo reader's interpretation of the outputs. Hence, we focus on the six distinct error patterns observed in this section.

Table \ref{tab:errors} shows the distribution of the stated errors in the transcripts from the outputs of the models. For non-Mizo-script outputs, SraVaani 1.0 in the zero-shot condition produced the highest number of affected utterances: $21$ sentences entirely in non-Mizo script. Of these $18$ were in the Meitei-mayek script, indicating that the system mistook the Mizo speech for Meitei-lon (Manipuri language). The remaining $3$ were generated in the Devanagari script, where the language is difficult to detect; however, they seem to approximate Hindi vocabulary. The Whisper-large V3 Mizo-fine-tuned model generated 4 outputs in Hebrew script mixed with Mizo script.

Overall, place names and people's names were substantially misrecognized. The highest number of misrecognitions occurred in the Sravaani 1.0 zero-shot model test, with 44 misrecognized name entities. However, when fine-tuned in the current work, the numbers were reduced to 24. For Whisper-fine-tuned models, the small and medium models misrecognized 17 and 18 named entities, respectively. However, with Whisper-large-v3 fine-tuning, the errors were reduced to 8. The reduction in named-entity errors after fine-tuning suggests that the training corpus provided useful exposure to the named-entity patterns present in the test set.

Glottal-stop recognition remained challenging across all evaluated systems. The SraVaani 1.0 model had the highest number of misrecognitions of the glottal stop, 9. However, fine-tuning in the current work reduced it to 6. The Whisper fine-tuned models showed the highest number of errors in recognising the glottal stop, i.e. 7. The number of misrecognitions decreased progressively as model capacity increased, with 6 misrecognitions in Whisper-medium and the lowest, 4, in the Whisper-large-v3 model.

SraVaani 1.0 also exhibited comparatively high error rates on Mizo numeral transcription. In total, there were 19 instances when the numerals were mistranscribed. However, when fine-tuned, the number reduced to only 2. Compared to that, the Mizo-fine-tuned Whisper models showed comparatively low numeral-transcription error rates, with three errors observed for Whisper-large-v3.

The SraVaani 1.0 model also had trouble when the Mizo speech contained code-mixed English words. There were far fewer errors in the Whisper systems. This may partly reflect the explicit multilingual coverage of English in the Whisper models. As seen in table \ref{tab:errors}, the errors in recognising code-mixed English words were 9, 3, and 2 for the Whisper small, medium, and large fine-tuned models. Fine-tuning reduced code-mixed English recognition errors in SraVaani from 49 to 19 occurrences. That shows that, apart from the Whisper models' coverage, the errors were lower in the fine-tuned Mizo models, as the Mizo data used for fine-tuning also included representative code-mixed English.

For the Mizo-specific orthographic distinction represented by $<$\textsubdot{t}$>$, SraVaani 1.0 exhibited 31 errors in the zero-shot condition. Following Mizo-specific fine-tuning, this number decreased to four. This may reflect limitations in the pretrained model's representation of this Mizo-specific phonological/orthographic distinction. In total, 31 $<$\textsubdot{t}$>$ were undetected in testing.

\section{Discussion and Conclusion}
\label{sec:discussion_conclusion}

This study presented a publicly available Mizo speech corpus and investigated the adaptation of multilingual pretrained ASR models to Mizo using a speaker-independent evaluation framework. The
experiments demonstrate that Whisper can be effectively adapted to Mizo even though the language is not included in the original Whisper multilingual language set. Among the three Whisper variants, performance improved consistently with increasing model capacity,
with Whisper-large-v3 achieving the lowest conventional WER of 18.08\%. SraVaani~1.0, despite supporting Mizo in its multilingual
pretrained model, produced substantially higher error rates in the zero-shot condition; however, Mizo-specific fine-tuning reduced its
WER from 58.27\% to 29.45\%, demonstrating the importance of language-specific adaptation. Across all evaluated systems, the morphology-aware Mizo WER was substantially lower than the conventional WER, indicating that a significant proportion of apparent word-level discrepancies is attributable to differences in the placement of morphological boundaries while preserving the same concatenated character sequence. These findings demonstrate the utility of large-scale multilingual pretraining and language-specific adaptation for low-resource Mizo ASR, while also demonstrating that standard WER alone may not fully reflect recognition quality in languages with variable morphological word boundaries. The publicly released corpus and the resulting ASR models provide a foundation for further development and benchmarking of Mizo speech technology.    

\section{Data availability and demonstrations}
This study reports $17.62$ hours of data, containing $7656$ sentences in the training set, $426$ sentences in the validation set, and $192$ sentences in the test set. Of these $5845$ sentences in the training set, the entire validation and test sets are available under the \textit{ANRF Open License for Software, Models and Datasets} at the AI-Kosh repository named IITG-Mizospeech-220726-V2. The remaining training data will be uploaded very soon. The repository can be accessed at: \blue{\href{https://aikosh.indiaai.gov.in/home/datasets/details/iitg_mizospeech_220726_v2.html}{https://aikosh.indiaai.gov.in}}. The models can be shared for research purposes on request.

Continuous Mizo speech recognition demonstrations utilizing the various Whisper fine-tuned models in this study are available at 
\blue{\href{https://clst.iitg.ac.in/tts/mizo-as/}{https://clst.iitg.ac.in/tts/mizo-as}}.

\section{Acknowledgment}
The funding for data collection in the current work was received from the Mizo Language Modelling Hackathon, organized by Digital India BHASHINI Division, Government of India and the Government of Mizoram. The computational resources are provided by the Center for Linguistic Science and Technology, Indian Institute of Technology Guwahati.

The authors would like to express their gratitude to Dr K. Samudravijaya, KL University; Dr Neeraj Kumar Sharma, IIT Guwahati; Dr Wendy Lalhminghlui, University of Bern and Ms Remruatpuii of HATIM, Mizoram, for their comments and suggestions.

\newpage
\bibliographystyle{IEEEtran}
\bibliography{mybib}

\end{document}